\documentclass[journal,transmag]{IEEEtran}

\usepackage{amsmath}
\usepackage{graphicx}
\usepackage{titlesec}
\usepackage{amsfonts,amssymb}
\usepackage{caption}
\usepackage{booktabs}
\usepackage{array}
\usepackage{cite}
\usepackage[numbers,sort&compress]{natbib}
\begin{document}

\title{Crossfusor: A Cross-Attention Transformer Enhanced Conditional Diffusion Model for Car-Following Trajectory Prediction}

\author{Junwei You,
        Haotian Shi*,
        Keshu Wu,
        Keke Long,
        Sicheng Fu,
        Sikai Chen*,
        Bin Ran
\thanks{*Corresponding authors: Haotian Shi (Email: hshi84@wisc.edu), Sikai Chen (Email: sikai.chen@wisc.edu)

Junwei You, Haotian Shi, Keshu Wu, Keke Long, Sicheng Fu, Sikai Chen, Bin Ran are with the Department of Civil and Environmental Engineering, University of Wisconsin-Madison, WI, USA.

Keshu Wu is with the Department of Landscape Architecture and Urban Planning, Texas A\&M University, TX, USA.}}



\maketitle

\begin{abstract}
\textit{Abstract}---Vehicle trajectory prediction is crucial for advancing autonomous driving and advanced driver assistance systems (ADAS), enhancing road safety and traffic efficiency. While traditional methods have laid foundational work, modern deep learning techniques, particularly transformer-based models and generative approaches, have significantly improved prediction accuracy by capturing complex and non-linear patterns in vehicle motion and traffic interactions. However, these models often overlook the detailed car-following behaviors and inter-vehicle interactions essential for real-world driving scenarios. This study introduces a Cross-Attention Transformer Enhanced Conditional Diffusion Model (Crossfusor) specifically designed for car-following trajectory prediction. Crossfusor integrates detailed inter-vehicular interactions and car-following dynamics into a robust diffusion framework, improving both the accuracy and realism of predicted trajectories. The model leverages a novel temporal feature encoding framework combining GRU, location-based attention mechanisms, and Fourier embedding to capture historical vehicle dynamics. It employs noise scaled by these encoded historical features in the forward diffusion process, and uses a cross-attention transformer to model intricate inter-vehicle dependencies in the reverse denoising process. Experimental results on the NGSIM dataset demonstrate that Crossfusor outperforms state-of-the-art models, particularly in long-term predictions, showcasing its potential for enhancing the predictive capabilities of autonomous driving systems.
\end{abstract}


\IEEEdisplaynontitleabstractindextext

\IEEEpeerreviewmaketitle

\section{Introduction}
\IEEEPARstart{V}{ehicle} trajectory prediction is a critical component in the advancement of autonomous driving and advanced driver assistance systems (ADAS), enabling informed decision-making that enhances road safety and traffic efficiency. The ability to forecast future vehicle positions is essential for collision avoidance, route planning, and adaptive cruise control. Over the years, various methodologies have been developed, ranging from traditional physics-based approaches to modern deep learning techniques.

While traditional methods have laid the groundwork, deep learning techniques have significantly advanced the field by offering more robust and adaptive solutions. Deep learning methods leverage delicate architectures capable of learning complex, nonlinear patterns and temporal dependencies from large datasets, usually surpassing both physics-based and classic machine learning methods \cite{huang2022survey}. For instance, models based on Recurrent Neural Networks (RNNs), Long Short-Term Memory (LSTM), Gated Recurrent Units (GRUs) and their variants  \cite{altche2017lstm}, \cite{dai2019modeling}, \cite{benterki2019long}, \cite{long2024physics}, capture long-term dependencies and non-linearity in vehicle motion, offering improvements in handling diverse traffic conditions. Convolutional Neural Networks (CNNs) and its variants that usually combine LSTM structures \cite{nikhil2018convolutional}, \cite{xie2020motion}, \cite{mo2020interaction}, \cite{wang2023vehicle}, process spatial-temporal data, effectively recognizing intricate patterns in vehicle trajectories. Transformer-based models \cite{chen2021s2tnet}, \cite{chen2022vehicle}, \cite{zhang2022ai}, \cite{geng2023physics}, excel in capturing interactions between multiple agents in dynamic traffic environments, providing state-of-the-art accuracy in trajectory prediction. Models based on Graph Neural Networks (GNNs) \cite{mo2021graph}, \cite{chen2021spatial}, \cite{lu2022vehicle} and Graph Convolutional Networks (GCN) \cite{sheng2022graph}, \cite{wu2023graph}, \cite{sadid2024dynamic}, enhance predictions by modeling relational dependencies between vehicles, enabling a deeper understanding of traffic flow and behavior. 

Additionally, generative models further enhance vehicle trajectory prediction by capturing the inherent uncertainty and variability in driving behavior that deterministic models often overlook. These models can generate a distribution of possible future trajectories, producing a more comprehensive and realistic prediction framework. The major types of generative models include Generative Adversarial Networks (GANs), flow-based methods, Variational Autoencoders (VAEs), and diffusion models. In general, GAN-based models utilize a generator to produce plausible trajectories and a discriminator to evaluate their realism, refining predictions through adversarial training. For example, the Driving Style Attention Generative Adversarial Network (DSA-GAN) \cite{choi2021dsa} incorporates driving style recognition through a Conditional GAN, which generates realistic multi-modal trajectories conditioned on different driving styles. The Conditional AutoEncoder Generative Adversarial Network (CAE-GAN) \cite{chen2022cae} is a hybrid model that combines Convolutional Autoencoders and GAN so as to capture complex spatiotemporal dependencies in traffic scenarios. The Multi-Agent Tensor Fusion GAN (MATF GAN) \cite{zhao2019multi} extracts interactions from historical trajectories and context based on convolutional fusion. A multi-vehicle collaborative learning model with spatio-temporal tensor fusion (TS-GAN) \cite{wang2020multi} models multi-agent spatial-temporal relations using an integrated generative adversarial framework. Unlike GANs, flow-based approaches transform a simple distribution into a complex one by learning invertible mappings, enabling the generation of diverse trajectories. For instance, a Diversity Sampling for Flow (DSF) technique \cite {ma2020diverse} is proposed to learn the sampling distribution that induces diverse and plausible trajectory predictions. The Mixed Gaussian Flow (MGF) model \cite{chen2024mixed} transforms a mixed Gaussian prior into the future trajectory manifold, which enhances the generation of diverse and controllable trajectory patterns. VAEs and their variants typically encode trajectories into a latent space and decode them back, allowing the generation of diverse trajectories by sampling from the latent space. Specifically, the recurrent VAE network \cite{de2022vehicles} leverages a conditional VAE to incorporate intention prediction and improve the accuracy and robustness of trajectory forecasts. The multi-scale VAE (Muse-VAE) \cite{lee2022muse} is designed for long-term trajectory prediction in environment-aware scenarios, which captures a multi-modal distribution of future trajectories influenced by various factors like vehicles and humans. Another intention-based conditional VAE model \cite{feng2019vehicle} leverages the probabilistic model to generate diverse and accurate future trajectories by incorporating driver intentions into the prediction process. The study \cite{neumeier2021variational} proposes a VAE model with an interpretable latent space for trajectory prediction, addressing the challenge of missing interpretability in traditional VAEs and improving the understanding of predicted trajectories. The research \cite{kim2021driving} combines driving style recognition with trajectory prediction using a conditional VAE, with the focus on in-vehicle CAN-bus data. 

Diffusion models have recently gained prominence in vehicle trajectory prediction due to their robust ability to handle uncertainty and generate diverse, realistic trajectories. Starting from a simple, usually Gaussian, distribution, diffusion models gradually transform this distribution into the complex distribution of real-world trajectories by learning the underlying data structure. Moreover, diffusion models have shown superiority in handling complex traffic scenarios where interactions between multiple agents and the environment need to be carefully considered. The ability and flexibility of the diffusion framework to incorporate spatial and temporal dependencies through advanced architectures, such as Transformer networks and GNNs, further enhance their predictive performance. In summary, diffusion models' capability to learn and generalize from large datasets, combined with their robustness to noise and uncertainty, makes them a prevailing method for predicting trajectories. There have been plenty of studies in literature that verified this. A diffusion-based model for environment-aware trajectory prediction is introduced in \cite{westny2024diffusion}, where its robustness and ability to handle complex traffic scenarios by leveraging conditional diffusion processes to model trajectory uncertainty is highlighted. A multi-modal vehicle trajectory prediction framework presented in \cite{li2023multi} uses a conditional diffusion model to address trajectory sparsity and irregularity in world coordinate systems. Combined with CNNs, a hierarchical vector transformer diffusion model developed in \cite{tang2024hierarchical} captures trajectory uncertainty and further enhances prediction performance. Another trajectory prediction framework called motion indeterminacy diffusion (MID) is introduced in \cite{gu2022stochastic}, which is designed to handle the indeterminacy of human behavior and provide accurate stochastic trajectory predictions. Recent study \cite{tang2024utilizing} has also explored the use of a diffusion model for pedestrian trajectory prediction in semi-open autonomous driving environments, focusing on reducing computational overhead and improving the accuracy of multi-agent joint trajectory predictions. The Conditional Equivariant Diffusion Model (EquiDiff) \cite{chen2023equidiff} combines the diffusion model with SO(2) equivariant transformer in order to utilize the geometric properties of location coordinates. It also applies RNNs and Graph Attention Networks (GAT) to extract social interactions from historical trajectories. 

Although the majority of diffusion-based trajectory prediction models, such as those analyzed above, have effectively incorporated social interactions across agents and environment into their structures for performance improvement, the integration of detailed car-following behaviors and complex vehicular interactions remains underexplored. Understanding the detailed, microscopic inter-vehicle interactions is essential as these factors are fundamental to real-world driving, where vehicles constantly respond to each other's specific movements to maintain safety and efficiency. Accurately modeling these dynamics in trajectory prediction is crucial for capturing the complexities of actual traffic, beyond general traffic interactions. However, latest models excel at modeling broad patterns and interactions in traffic, while they often overlook the finer details of car-following dynamics, such as how closely a vehicle follows another, and acceleration and deceleration patterns. 

In view of this, this study proposes a Cross-Attention Transformer Enhanced Conditional Diffusion Model (Crossfusor) for trajectory prediction, dedicated to bridging this gap by integrating detailed car-following behaviors into the model structure. Crossfusor leverages comprehensive temporal encoding techniques and attention mechanisms to capture the fine-grained interactions between vehicles. By incorporating these specific dynamics, the model not only improves the prediction accuracy but also enhances the realism of generated trajectories, reflecting the complex and varied nature of real-world driving. This approach addresses the limitations of current models and potentially offers an advancement in the predictive capabilities of autonomous and assisted driving systems. The main contributions of this paper are threefold:
\begin{itemize}
\item We develop a novel temporal feature encoding framework consisting of GRU, vehicle location-based attention mechanism, and Fourier embedding which effectively extracts the temporal features from the historical vehicle trajectory.

\item We propose to use the noise scaled by the encoded historical feature to replace the isotropic Gaussian noise in the diffusion model, which introduces an oriented forward noise addition process informed by history.

\item We integrate the diffusion model with a cross-attention transformer-based architecture which thoroughly models the intricate car-following dependencies and dynamic inter-vehicle interactions, aiming at guiding the reverse denoising process and directing the trajectory generation from noise. 
\end{itemize}

The remainder of this paper is organized as follows: Section \uppercase\expandafter{\romannumeral2} systematically formulate the problem. Section \uppercase\expandafter{\romannumeral3} presents the methodology employed in this study. Experimental results and corresponding observations are discussed in Section \uppercase\expandafter{\romannumeral4}, and Section \uppercase\expandafter{\romannumeral5} concludes the paper. 

\section{Problem Description of Trajectory Prediction in Car-Following Dynamics}
The complex nature of the driving environment requires a thorough understanding of the vehicle's current and historical states, as well as the dynamic interactions between neighboring vehicles. This study introduces a novel diffusion model for predicting the future trajectory of vehicles in a car-following scenario, specifically focusing on a three-vehicle platoon: a leading vehicle, a study vehicle, and a following vehicle. Unlike existing models, the proposed Crossfusor model aims at capturing the intricate dynamics and the probabilistic nature of inter-vehicular dependencies, utilizing historical trajectory data, speed profiles, and inter-vehicle spacing. 

To formalize the problem, let $\mathbf{x}_i^{\rm his}$ and $\mathbf{v}_i^{\rm his}$ denote the historical trajectory and speed of the $i$-th vehicle in a three-vehicle platoon at time, where $i=\rm stu$ for the study vehicle, $i=\rm lea$ for the leading vehicle of the study vehicle, and $i=\rm fol$ for the following vehicle of the study vehicle. $\Delta\mathbf{x}_1^{\rm his}$ and $\Delta\mathbf{x}_2^{\rm his}$ represent the historical spacing between the leading and study vehicles, and that between the study and following vehicles, respectively. Hence, the goal is to predict the future trajectory $\hat{\mathbf{x}}_{\rm stu}^{\rm fut}$ of the study vehicle given the historical information stated above, as shown in the following equation:
\begin{equation}
    \hat{\mathbf{x}}_{\rm stu}^{\rm fut} = f(\mathbf{x}_{\rm stu}^{\rm his}, \mathbf{v}_{\rm stu}^{\rm his}, \mathbf{x}_{\rm lea}^{\rm his}, \mathbf{v}_{\rm lea}^{\rm his}, \mathbf{x}_{\rm fol}^{\rm his}, \mathbf{v}_{\rm fol}^{\rm his}, \Delta\mathbf{x}_1^{\rm his}, \Delta\mathbf{x}_2^{\rm his})
\end{equation}
where $f(\cdot)$ embodies the proposed Crossfusor model.

\section{Methodology}
\subsection{Overview}
\subsubsection{Denoising Diffusion Probabilistic Models}
Denoising Diffusion Probabilistic Models (DDPMs) \cite{ho2020denoising}, widely recognized as diffusion models, are a class of generative models that have garnered attention for their ability to synthesize high-quality samples across various domains. At their core, DDPMs operate by simulating a forward diffusion process gradually adding noise to the data over a series of time steps, which ends up transforming the original data distribution into a Gaussian distribution. This process is parameterized by a Markov chain where each step is conditioned on the previous one, ensuring a smooth transition from data to noise. The essence of DDPMs lies in their reverse process, a learned denoising process that aims to reconstruct the original data from the noise. This reverse process can be modeled by a deep neural network to estimate the parameters of the Gaussian distribution at each step, gradually removing noise to recover the clean data. By carefully designing this network, DDPMs can generate novel data samples that closely mimic the characteristics of the original dataset. This capability has profound implications for trajectory prediction, as it enables the generation of plausible future states based on observed data, offering a powerful tool for forecasting and simulation in dynamic systems. 
\subsubsection{Overall Model Framework}
Taking advantage of diffusion models, the overall framework of the proposed Crossfusor for predicting the future trajectory of a vehicle within a platoon, utilizing historical driving data and detailed vehicular interactions, is shown in Figure 1. 
\begin{figure*}[htbp]
\centering
\includegraphics[width=14cm]{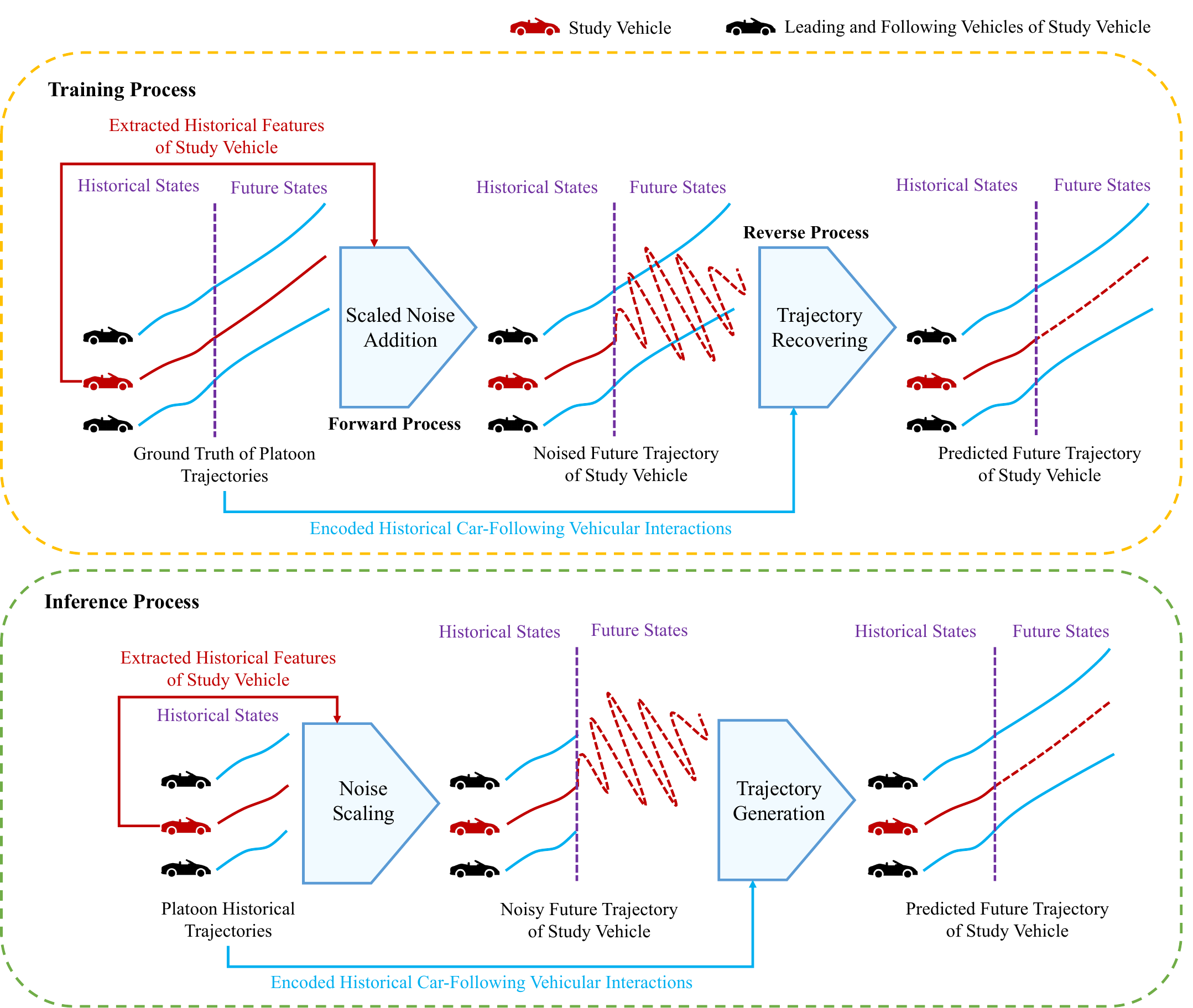}
\caption{Overall Framework of Crossfusor}
\label{fig:yourlabel}
\end{figure*}
Specifically, in the training process, historical features of the study vehicle are extracted to scale the noise added throughout the forward process, while the comprehensive car-following interactions are then encoded and supplemented to a denoising network in the reverse process for trajectory reconstruction from the completely noisy state. The inference process mirrors the training process but gets rid of the forward process. Starting with a random noise scaled by the extracted historical features, the encoded vehicular interactions will guide the reverse denoising process to generate the future trajectory of the study vehicle from the initially scaled noise. In the following sections, the components in both the forward and reverse processes of Crossfusor will be elaborated on in detail.
\subsection{Forward Process}
\subsubsection{Historical Feature Extraction}
As addressed above, in the context of trajectory prediction, the forward process normally stands for adding noise upon future trajectory $\mathbf{x}_{\rm stu}^{\rm fut}$ successively. In the innovative approach presented in this paper, the forward process diverges from the traditional methodology. Specifically, adding noise to the future trajectory is not a purely stochastic process but is scaled by historical features extracted through a specialized procedure. By conditioning the noise on historical features, we ensure that the forward diffusion process incorporates necessary conditions or restrictions reflective of the system's true dynamics. This results in a more informed and directed process of transitioning from data to noise, ensuring that the generated future trajectories are not only a product of random noise but are informed by the system's past. In this context, extracting meaningful features from historical trajectory remains critical. The overall structure of the forward process is illustrated in Figure 2, where the historical feature of the study vehicle is abstracted from historical trajectory as an encoded sequence $\mathbf{z}_{\rm stu}^{\rm his}$, through a specially designed pipeline, and is further represented as a global value termed virtual variance which is integrated into the noise at each time step to create the scaled noise. 
\begin{figure}
\centering
\includegraphics[width=9cm]{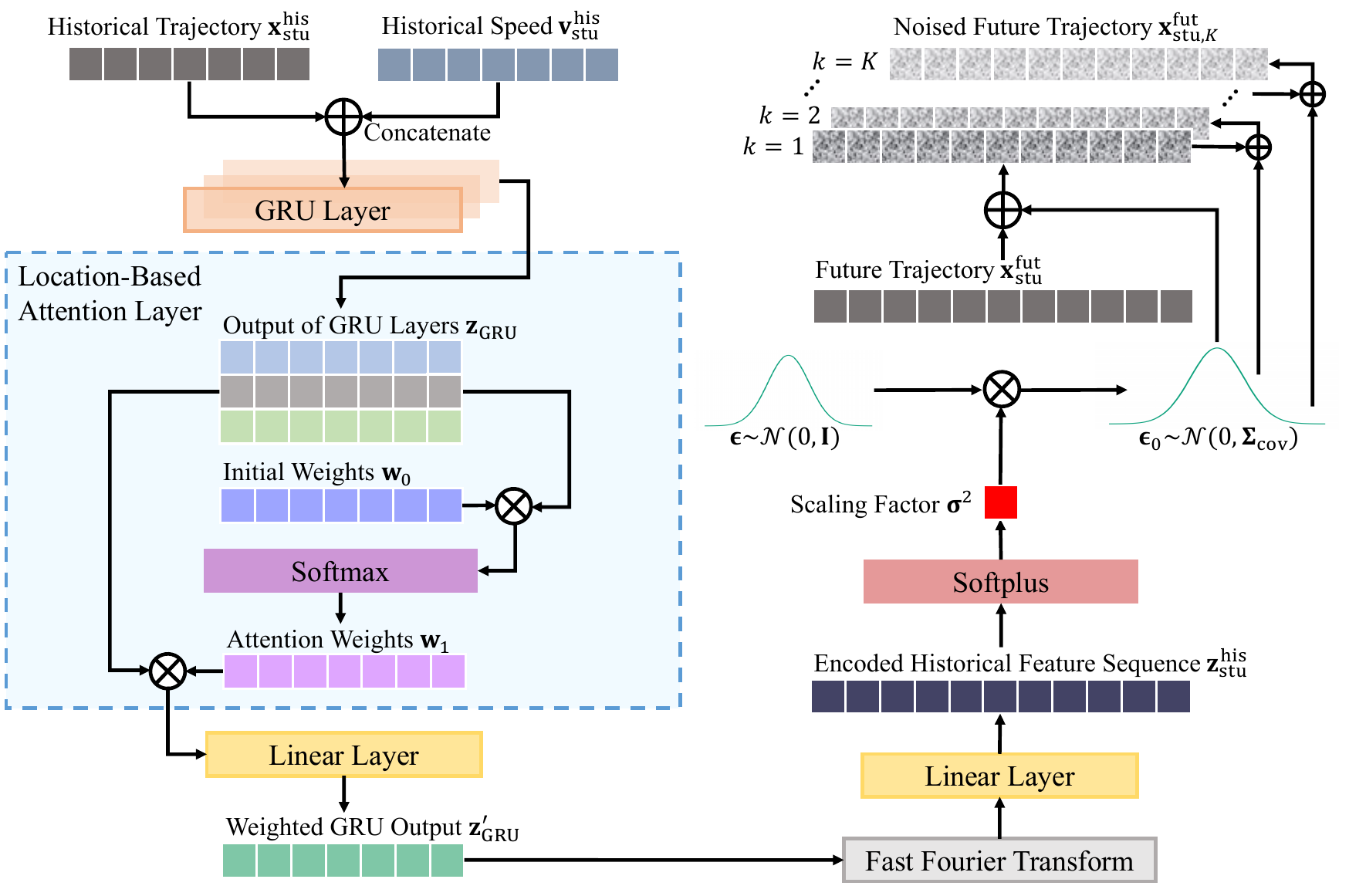}
\caption{Structure of Historical Trajectory Feature Extraction and Forward Process}
\label{fig:yourlabel}
\end{figure}

Concretely, historical trajectory $\mathbf{x}_{\rm stu}^{\rm his}$ is concatenated with historical speed $\mathbf{v}_{\rm stu}^{\rm his}$, and then fed into stacked GRU layers \cite{dey2017gate} for data fusion and temporal feature abstraction. The output of GRU layers denoted as $\mathbf{z}_{\rm GRU}$ is then passed through a location-based attention layer. Location-based attention is a mechanism that allows a model to focus on specific parts of an input sequence when generating an output. Practically, it is implemented by assigning weights to different positions or locations within the input sequence. These weights are then used to modulate the model's focus, allowing it to focus more on certain areas based on their location. This mechanism enables the model to capture sequential dependencies and spatial relationships within the data. In our case, it allows the model to focus on specific and key trajectory segments based on the vehicle's spatial and temporal location. Given the initial attention weights $\mathbf{w}_0$, the operation of the location-based attention mechanism is formulated as follows:
\begin{equation} 
    \mathbf{w}_1 = \text{softmax}(\mathbf{W} \cdot (\mathbf{z}_{\rm GRU} \odot \mathbf{w}_0)+\mathbf{b})
\end{equation}
\begin{equation} 
    \mathbf{z}_{\rm loc} =  \mathbf{w}_1 \odot \mathbf{z}_{\rm GRU}
\end{equation}
where $\mathbf{W}$ is the weight matrix for linear projection, $\mathbf{b}$ is a bias vector, $\mathbf{w}_1$ represents the updated attention weights, $\mathbf{z}_{\rm loc}$ stands for the output of location-based attention layer, and $\odot$ refers to the operation of Hadamard product. Subsequently, after another linear projection, the GRU output $\mathbf{z}_{\rm GRU}$ is transformed into a weighted sequence denoted as $\mathbf{z}_{\rm GRU}^{'}$. 

While GRU is adept at capturing long-term temporal dependencies of a sequence, another widely used algorithm for time series encoding, known as Fast Fourier transform (FFT) \cite{duhamel1990fast}, \cite{zhang2023learnable}, excels at representing a sequence by decomposing it into its constituent frequencies so as to reveal the periodic patterns. Taking the weighted GRU output $\mathbf{z}_{\rm GRU}^{'}$ as input, FFT is formulated as the equation below:
\begin{equation} 
    \mathbf{z}_{\rm FFT}[i] = \sum_{n=0}^{N-1} \mathbf{z}_{\rm GRU}^{'}[n] \cdot e^{-j \cdot 2 \pi \cdot \frac{i \cdot n}{N}}
\end{equation}
where $N$ is the length of the input sequence, $\mathbf{z}_{\rm GRU}^{'}[n]$ is the value of the input time-domain sequence at the $n$-th sample, $n \in [0,N-1]$, $e^{-j \cdot 2 \pi \cdot \frac{i \cdot n}{N}}$ is the complex exponential function that represents the basis functions of FFT, $j$ is the imaginary unit, and $\mathbf{z}_{\rm FFT}[i]$ is the value of the transformed frequency-domain sequence at the $i$-th frequency bin, $i \in [0,N-1]$. A linear layer is further applied, which finally yields the encoded historical feature $\mathbf{z}_{\rm stu}^{\rm his}$ with temporal dependencies and patterns effectively extracted. 
\subsubsection{Scaled Noise Addition}
As mentioned previously, in this study, we propose to add the noise scaled by the extracted historical trajectory features in each time step during the forward process. Statistically, this means that instead of sampling from the standard normal distribution $\mathcal{N}(0, \mathbf{I})$ as what traditional diffusion models would do, noise now is sampled from the normal distribution $\mathcal{N}(0, \mathbf{\Sigma}_{\rm cov})$, where the covariance matrix $\mathbf{\Sigma}_{\rm cov}$ is a diagonal matrix, and is represented from the encoded historical trajectory $\mathbf{z}_{\rm stu}^{\rm his}$. Specifically, to find  $\mathbf{\Sigma}_{\rm cov}$, we first take the mean of each dimension of $\mathbf{z}_{\rm stu}^{\rm his}$ which yields a vector denoted as $\mathbf{\mu}$, and then apply the Softplus activation function \cite{zheng2015improving} upon $\mathbf{\mu}$. The resulting vector, denoted as $\mathbf{\sigma}^2$, is used as the scaling factor to reshape the standard normal distribution to maintain the variance of $\mathbf{\Sigma}_{\rm cov}$. This process can be formulated as follows:
\begin{equation} 
    \mathbf{\sigma}^2 = \log(1 + e^{\mathbf{\mu}})
\end{equation}
\begin{equation}
    \mathbf{\Sigma}_{\rm cov} = \mathbf{\sigma}^2\mathbf{I}
\end{equation}

In practice, given that $\mathbf{\epsilon}_0 \sim \mathcal{N}(0, \mathbf{I})$ is an independent standard normal variable randomly sampled from a standard normal distribution, the scaled noise denoted as $\mathbf{\epsilon}$ can also be expressed and calculated directly as follows:
\begin{equation}
   \mathbf{\epsilon} = \mathbf{\Sigma}_{\rm cov}^{\frac{1}{2}}\mathbf{\epsilon}_0 = \text{diag}(\mathbf{\sigma}) \cdot \mathbf{\epsilon}_0, \quad \mathbf{\epsilon} \sim \mathcal{N}(0, \mathbf{\Sigma_{\rm cov}})
\end{equation}
where $\text{diag}(\mathbf{\sigma})$ refers to a diagonal matrix where each element of the vector $\mathbf{\sigma}$ is placed on the main diagonal of the matrix.

On this basis, the forward incremental noise addition process will take the future trajectory $\mathbf{x}^{\rm fut}_{\rm stu}$ as input and gradually add the said scaled noise to the input for $K$ time steps, which is formulated as follows:
\begin{equation}
    \mathbf{x}_{{\rm stu},k}^{\rm fut} = \sqrt{\alpha_k} \mathbf{x}_{{\rm stu},k-1}^{\rm fut} + \sqrt{\beta_k} \mathbf{\epsilon}, \quad \mathbf{\epsilon} \sim \mathcal{N}(0, \mathbf{\Sigma}_{\rm cov})
\end{equation}
\begin{equation}
    q(\mathbf{x}_{{\rm stu},k}^{\rm fut} | \mathbf{x}_{{\rm stu},k-1}^{\rm fut}) = \mathcal{N}(\mathbf{x}_{{\rm stu},k}^{\rm fut}; \sqrt{\alpha_k} \mathbf{x}_{{\rm stu},k-1}^{\rm fut}, \beta_k \mathbf{\Sigma}_{\rm cov})
\end{equation}
where $\beta_k$ is the time step-specific factor to control the intensity of the noise added at each step, $\mathbf{\epsilon}$ represents the noise vector sampled from a Gaussian distribution with covariance matrix $\mathbf{\Sigma}_{\rm cov}$ as stated above, $\mathbf{x}_{{\rm stu},k}^{\rm fut}$ is the data distribution at time step $k$ after undergoing $k$ times of noise addition, $\mathbf{x}_{{\rm stu},k-1}^{\rm fut}$ is the data vector at the previous time step $k-1$, and $\alpha_k = 1 - \beta_k$. 

Define $\bar{\alpha}_k = \prod_{i=1}^{k} \alpha_i$, and we can derive the diffusion process at any step $k$ from the original data $\mathbf{x}_{\rm stu}^{\rm fut}$ in a closed form:
\begin{equation}
\mathbf{x}_{{\rm stu},k}^{\rm fut} = \sqrt{\bar{\alpha}_k} \mathbf{x}_{\rm stu}^{\rm fut} + \sqrt{(1 - \bar{\alpha}_k)} \mathbf{\epsilon}, \quad \mathbf{\epsilon} \sim \mathcal{N}(0, \mathbf{\Sigma}_{\rm cov})
\end{equation}
\begin{equation}
q(\mathbf{x}_{{\rm stu},k}^{\rm fut} | \mathbf{x}_{\rm stu}^{\rm fut}) = \mathcal{N}(\mathbf{x}_{{\rm stu},k}^{\rm fut}; \sqrt{\bar{\alpha}_k} \mathbf{x}_{\rm stu}^{\rm fut}, (1 - \bar{\alpha}_k) \mathbf{\Sigma}_{\rm cov})
\end{equation}
Ultimately, when  $K \xrightarrow{} \infty$, $\mathbf{x}_{{\rm stu},K}^{\rm fut}$ will approximate to follow the prior noise distribution used in the diffusion process, $\mathbf{x}_{{\rm stu},K}^{\rm fut} \sim \mathcal{N}(0, \mathbf{\Sigma}_{\rm cov})$. The distribution of the entire sequence from $\mathbf{x}_{{\rm stu}}^{\rm fut}$ to $\mathbf{x}_{{\rm stu},K}^{\rm fut}$ conditioned on the original data $\mathbf{x}_{{\rm stu}}^{\rm fut}$ is shows as follows:
\begin{equation}
    q(\mathbf{x}_{{\rm stu},1:K}^{\rm fut} |  \mathbf{x}_{{\rm stu}}^{\rm fut}) = \prod_{k=1}^{K} q( \mathbf{x}_{{\rm stu},k}^{\rm fut} |  \mathbf{x}_{{\rm stu},k-1}^{\rm fut})
\end{equation}

\subsection{Reverse Process}
\subsubsection{Car-Following Vehicular Interaction Encoding}
The reverse process aims at generating future trajectory $\hat{\mathbf{x}}_{\rm stu}^{\rm fut}$ from the corresponding historical information informed noise $\mathbf{\epsilon} \sim \mathcal{N}(0, \mathbf{\Sigma}_{\rm cov})$. In the reverse process, the denoising network plays a crucial role in precisely recovering and predicting future trajectories, taking advantage of historical information and microscopic adjacent vehicular interactions. Specifically, in this study, focusing on the three-vehicle platoon scenario, as addressed in the previous section, the historical trajectories and speed profiles of both the leading and following vehicles of the subject vehicle, as well as the inter-vehicle spacing, will be thoroughly encoded and embedded in the denoising network to guide the successive denoising process. 

In this case, effectively encoding the complex car-following variables becomes pivotal. Hence, we design a cross-attention transformer-based architecture \cite{gheini2021cross} dedicated to modeling the intricate dependencies and dynamic interactions between these variables. The entire process is demonstrated in Figure 3. 
\begin{figure}
\centering
\includegraphics[width=9cm]{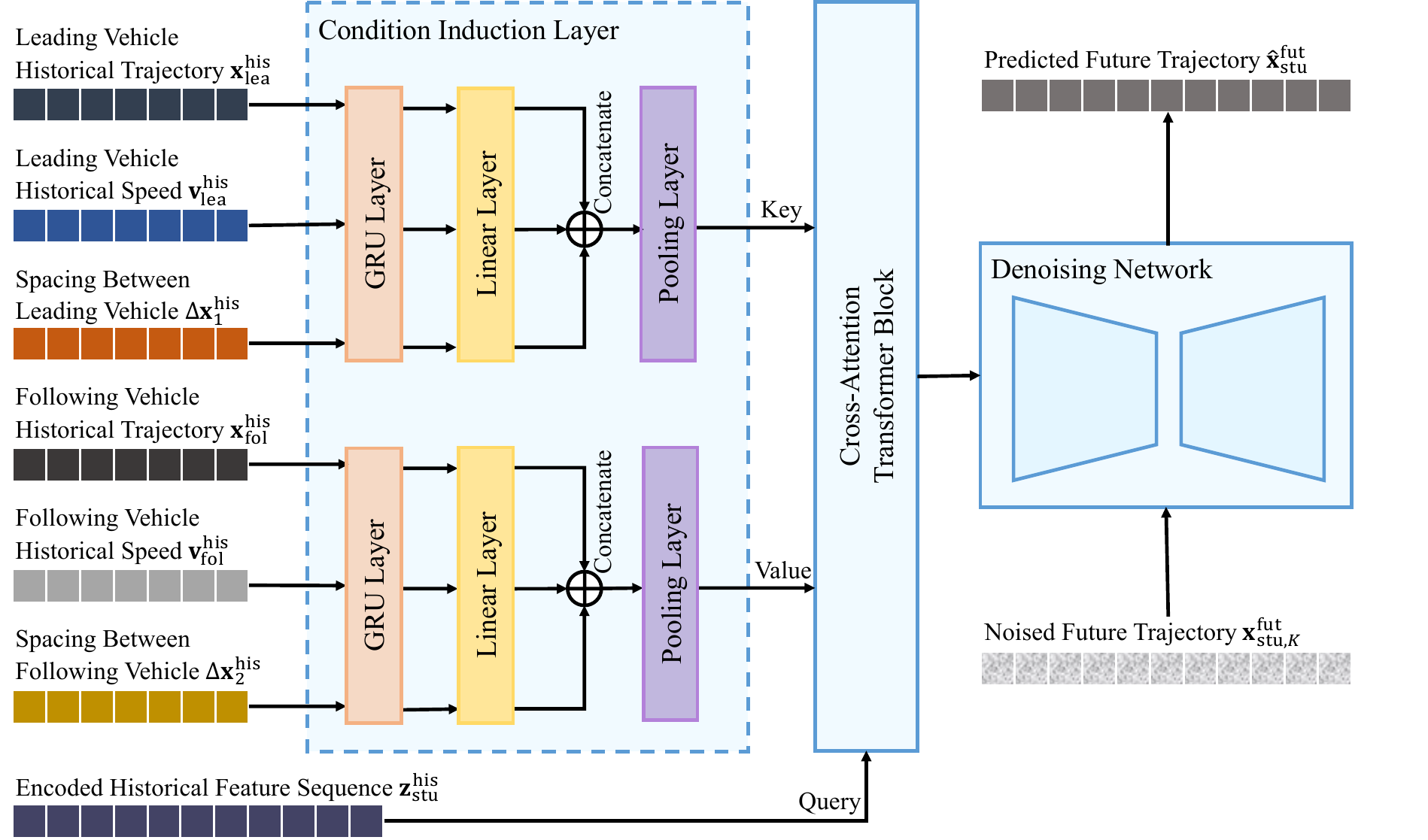}
\caption{Architecture of Reverse Process}
\label{fig:yourlabel}
\end{figure}
Digging into details, historical trajectories, speeds, and spacing for both leading and following vehicles are first processed by GRU layers to capture temporal dynamics. The extracted features are then linearized and concatenated to pass through a pooling layer, which forms the key and value vectors, denoted as $\mathbf{K}$ and $\mathbf{V}$, for the leading and following vehicles, respectively, where $\mathbf{K} \in \mathbb{R}^{d_k}$, $\mathbf{V} \in \mathbb{R}^{d_v}$. We can also treat the study vehicle's encoded trajectory $\mathbf{z}_{\rm stu}^{\rm his}$ as  query vector $\mathbf{Q}$, $\mathbf{Q} \in \mathbb{R}^{d_q}$. $\mathbf{Q}$ is used within the cross-attention transformer block to selectively weigh the leading and following vehicles' features, synthesizing a contextualized output that encapsulates the interactive behavior of the vehicles in a platoon. The formulation of $\mathbf{Q}$, $\mathbf{K}$, and $\mathbf{V}$ is shown as follows:
\begin{equation}
\mathbf{Q} = \mathbf{z}_{\rm stu}^{\rm his}
\end{equation}
\begin{equation}
    \begin{split}
        \mathbf{K} = \text{Pooling} (\text{Concat} (
        & \text{Linear}(\text{GRU}(\mathbf{x}_{\rm lea}^{\rm his})), \\
        & \text{Linear}(\text{GRU}(\mathbf{v}_{\rm lea}^{\rm his})), \\
        & \text{Linear}(\text{GRU}(\Delta \mathbf{x}_1^{\rm his})))
    \end{split}
\end{equation}
\begin{equation}
    \begin{split}
        \mathbf{V} = \text{Pooling} (\text{Concat} (
        & \text{Linear}(\text{GRU}(\mathbf{x}_{\rm fol}^{\rm his})), \\
        & \text{Linear}(\text{GRU}(\mathbf{v}_{\rm fol}^{\rm his})), \\
        & \text{Linear}(\text{GRU}(\Delta \mathbf{x}_2^{\rm his})))
    \end{split}
\end{equation}

The structure of the cross-attention transformer block is shown in Figure 4, where the multi-head cross-attention operation can be formulated as follows:
\begin{equation}
    \mathbf{z}_{\rm MCA} = \text{Concat}(\mathbf{head}_1, ..., \mathbf{head}_i, ..., \mathbf{head}_h) \mathbf{W}^{\rm out}
\end{equation}
where $\mathbf{W}^{\rm out}$ is the output weight matrix that linearly transforms the concatenated vector from all the heads into the desired output dimension, and each attention head, $\mathbf{head}_i$, is computed using the equation below:
\begin{equation}
    \mathbf{head}_i = \text{softmax}\left(\frac{\mathbf{Q}\mathbf{W}_i^{\rm que} (\mathbf{K}\mathbf{W}_i^{\rm key})^{\rm T}}{\sqrt{d_k}}\right) \cdot (\mathbf{W}_i^{\rm val})
\end{equation}
where $\mathbf{W}_i^{\rm que}$, $\mathbf{W}_i^{\rm key}$, $\mathbf{W}_i^{\rm val}$ are the parameter matrices specific to each head for the queries, keys, and values, respectively.
\begin{figure}
    \centering
    \includegraphics[width=9cm]{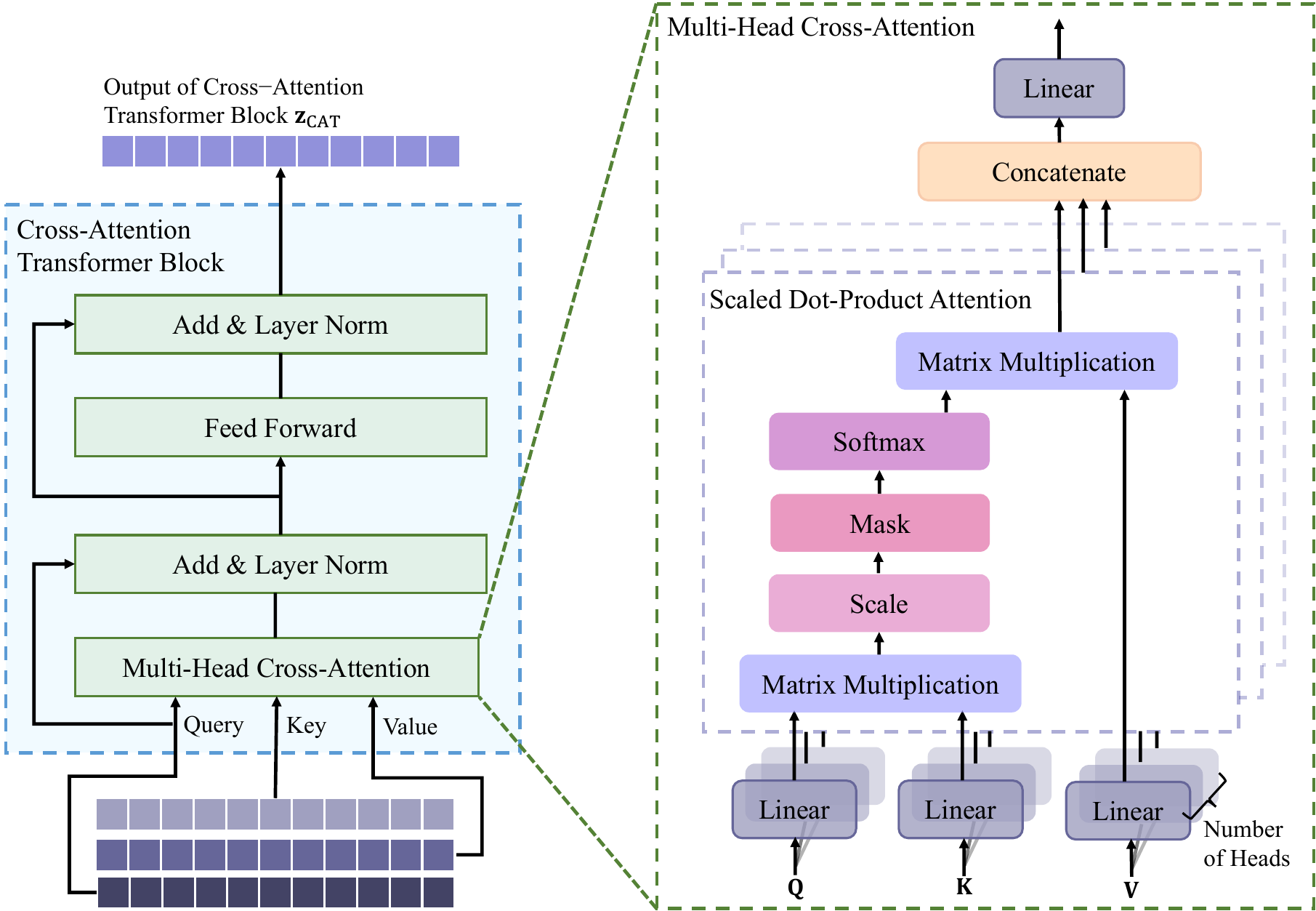}
    \caption{Structure of Cross-Attention Transformer Block}
    \label{fig:fig4}
\end{figure}

Finally, the output of the cross-attention transformer block enriched with vehicle relational information is embedded into a denoising network, directing and enhancing the prediction of the study vehicle's future trajectory. 
\subsubsection{Denoising Process}
Taking a scaled noise $\mathbf{\epsilon} \sim \mathcal{N}(0, \mathbf{\Sigma}_{\rm cov})$ as input, the denoising network reconstructs the corresponding clean and precise future trajectory of the study vehicle. In doing so, it reverses the diffusion process by sequentially predicting and removing the noise distribution introduced at each time step, thereby progressively restoring the corresponding trajectory to its original uncorrupted distribution. Given the estimated data distribution $\mathbf{x}_{{\rm stu},k}^{\rm fut}$ at any time step $k$ and the contextual information $\mathbf{c}$ from cross-attention block, estimation of the data distribution at time $k-1$ is shown as the following equation:
\begin{equation}
    p_\theta(\mathbf{x}_{{\rm stu},k-1}^{\rm fut}| \mathbf{x}_{{\rm stu},k}^{\rm fut}, \mathbf{c}) = \mathcal{N}\left(\mathbf{x}_{{\rm stu},k-1}^{\rm fut} ; \mathbf{\mu}_\theta(\mathbf{x}_{{\rm stu},k}^{\rm fut}, k, \mathbf{c}), \mathbf{\Sigma}_\theta(k)\right)
\end{equation}
where $\mathbf{\mu}_\theta(\mathbf{x}_{{\rm stu},k}^{\rm fut}, k, \mathbf{c})$ is the predicted mean for recovering $\mathbf{x}_{{\rm stu},k-1}^{\rm fut}$, informed by the context encoding $\mathbf{c}$, and $\mathbf{\Sigma}_\theta(k)$ is the learned covariance matrix at time step $k$. The joint probability over the sequence, conditioned on  $\mathbf{c}$, is given by:
\begin{equation}
    p_\theta(\mathbf{x}_{{\rm stu},0:K}^{\rm fut} | \mathbf{c}) = p(\mathbf{x}_{{\rm stu},K}^{\rm fut}) \prod_{k=1}^{K} p_\theta(\mathbf{x}_{{\rm stu},k-1}^{\rm fut} | \mathbf{x}_{{\rm stu},k}^{\rm fut}, \mathbf{c})
\end{equation}
\begin{equation}
    p(\mathbf{x}_{{\rm stu},K}^{\rm fut}) = \mathcal{N}(\mathbf{x}_{{\rm stu},K}^{\rm fut}; 0, \mathbf{\Sigma}_{\rm cov})
\end{equation}
\subsection{Training Objective}
The training objective of Crossfusor is to maximize the variational lower bound (ELBO). Incorporating the additional context from the cross-attention transformer block, the adapted ELBO is formulated as follows:
\begin{equation}
\begin{split}
    \mathcal{L}(\theta) & = \text{KL}(q(\mathbf{x}_{{\rm stu},K}^{\rm fut}|\mathbf{x}_{{\rm stu}}^{\rm fut} \Vert p(\mathbf{x}_{{\rm stu},K}^{\rm fut}))  -\log p_{\theta}(\mathbf{x}_{{\rm stu}}^{\rm fut} | \mathbf{x}_{{\rm stu},1}^{\rm fut}, \mathbf{c})\\
    & + \sum_{k=2}^{K} \text{KL}(q(\mathbf{x}_{{\rm stu},k-1}^{\rm fut} | \mathbf{x}_{{\rm stu},k}^{\rm fut}, \mathbf{x}_{{\rm stu}}^{\rm fut}) \Vert p_\theta(\mathbf{x}_{{\rm stu},k-1}^{\rm fut} | \mathbf{x}_{{\rm stu},k}^{\rm fut}, \mathbf{c}))
    \end{split}
\end{equation}
Moreover, the objective function can be simplified for a more stable training process. In particular, the expected negative log-likelihood can be replaced with a reconstruction loss such as the mean squared error (MSE) between the predicted noise and the true noise. The KL divergence term can be omitted. This will result in a simplified training objective function according to \cite{ho2020denoising}, as shown below:
\begin{equation}
    \mathcal{L}_{simp}(\theta) = \mathbb{E}_{\mathbf{x}_{{\rm stu}}^{\rm fut}, \mathbf{\epsilon}, k}\left[\| \mathbf{\epsilon} - \hat{\mathbf{\epsilon}}_\theta(\mathbf{x}_{{\rm stu},k}^{\rm fut}, k, \mathbf{c}) \|^2\right]
\end{equation}
The objective essentially focuses on the accuracy of noise prediction. In the denoising network, in this study, the noise at each time step is predicted through an adapted U-Net \cite{ronneberger2015u} that incorporates inter-vehicle car-following interactions within a platoon encoded by the cross-attention transformer block. The detailed architecture of the noise prediction network is illustrated in Figure 5. 
\begin{figure}[htbp]
\includegraphics[width=9cm]{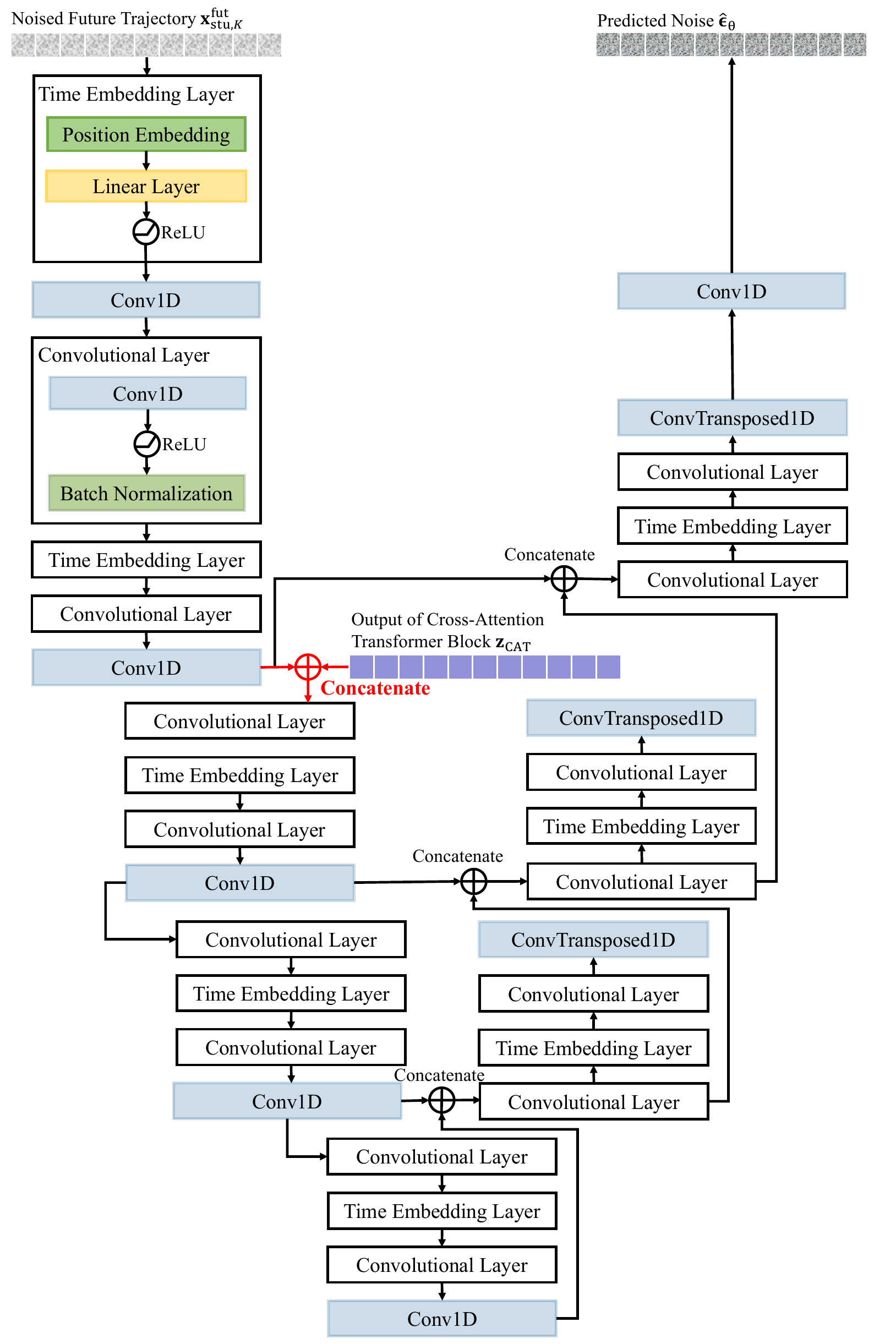}
\caption{Adapted U-Net for Noise Prediction}
\label{fig:yourlabel}
\end{figure}
\section{Experiment}
\subsection{Dataset}
Experiments are conducted to validate the performance of the proposed Crossfusor. The experiments are conducted on the Next Generation Simulation (NGSIM) dataset \cite{colyar2007us}, \cite{colyar2006us}. The NGSIM dataset stands as a rich repository of vehicle trajectory information, documenting traffic patterns along the eastbound I-80 in the San Francisco Bay area and the southbound US 101 in Los Angeles. This dataset offers a vivid portrayal of real-world highway scenarios. Through the lens of overhead camera recordings sampled at 10 Hz, it provides an intricate understanding of vehicular movements, serving as a cornerstone for in-depth analysis and modeling of traffic behavior.

After pre-processing, a selection of 1,606 three-vehicle platoon trajectory data is made carefully, totaling 4,818 individual trajectories. Each trajectory has a equal length of 200 frames. In this study, historical data spanning three seconds (30 frames) is utilized to predict a future horizon of five seconds (50 frames). Iterations is conducted for each group of platoon trajectories at one-second intervals. This process culminated in the generation of 192,480 sets of platoon trajectory data, corresponding to 577,440 individual vehicle trajectories. 173,232 sets of platoon trajectories are used for training purposes, while the remaining 19,248 sets are testing sets.
\subsection{Model Setups}
The proposed Crossfusor is developed based on PyTorch framework, and is trained on a single NVIDIA RTX 4090 GPU. AdamW algorithm is applied as the optimizer with an initial learning rate of 0.001. The training process converges substantially within 10 epochs with the batch size of 64. Detailed hyperparameters according the Crossfusor network architecture are shown in Table \uppercase\expandafter{\romannumeral1}. 
\begin{table}[ht]
  \centering
  \caption{Model Hyperparameters}
  \begin{tabular}{>{\centering}m{6cm} c}
    \toprule[1.0pt]
    Parameter & Value \\
    \midrule
    Hidden Size of GRU  & 50 \\
    No. of GRU Layers   & 2 \\
    Embedding Size of Cross-Attention Transformer & 50 \\
    No. of Attention Heads & 5 \\
    Feed Forward Size & 100 \\
    No. of Down Sampling Channels in U-Net & (8, 16, 32, 64, 128) \\
    No. of Up Sampling Channels in U-Net & (128, 64, 32, 16, 8) \\
    $K$ & 200 \\
    $\beta_0$ & 0.0001 \\
    $\beta_K$ & 0.02 \\
    \bottomrule[1.0pt]
  \end{tabular}
\end{table}
\subsection{Evaluation Metrics}
To evaluate the performance of Crossfusor developed in this study, the root of mean squared error (RMSE) of the predicted future trajectory $\hat{\mathbf{x}}_{\rm stu}^{\rm fut}$ and the ground truth $\mathbf{x}_{\rm stu}^{\rm fut}$is calculated, as the equation below:
\begin{equation}
    \text{RMSE} = \sqrt{\frac{1}{\Delta{t}} \sum_{i=t}^{t+\Delta{t}} (\mathbf{x}_{\rm stu}^{\rm fut}\big|_i  - \hat{\mathbf{x}}_{\rm stu}^{\rm fut}\big|_i)^2}
\end{equation}
where the future trajectory starts at time $t$ and $\Delta{t}$ is the future prediction horizon.

In addition, we also calculate the Final Displacement Error (FDE) and Average Displacement Error (ADE), as shown in the following equations:
\begin{equation}
    \text{FDE} = \left\| \mathbf{x}_{\rm stu}^{\rm fut}\big|_{t+\Delta{t}} - \hat{\mathbf{x}}_{\rm stu}^{\rm fut}\big|_{t+\Delta{t}} \right\|_2
\end{equation}
\begin{equation}
\text{ADE} = \frac{1}{\Delta{t}} \sum_{i=t}^{t+\Delta{t}} \left\| \mathbf{x}_{\rm stu}^{\rm fut}\big|_i - \hat{\mathbf{x}}_{\rm stu}^{\rm fut}\big|_i \right\|_2
\end{equation}

\subsection{Baseline Models}
In light of one of the latest work \cite{chen2023equidiff} that uses a diffusion model for trajectory prediction, we compare the proposed Crossfusor with the representative state-of-the-art generative trajectory prediction models evaluated in \cite{chen2023equidiff}, as listed below.  
\begin{itemize}
\item[$\bullet$] Constant Velocity (CV): CV uses a constant Kalman filter to forecast future positions, providing a basic yet effective benchmark for more sophisticated models.
\end{itemize}
\begin{itemize}
\item[$\bullet$] Vanilla LSTM (V-LSTM): V-LSTM uses a basic Long Short-Term Memory (LSTM) architecture to process historical trajectory data. This model excels at predicting future positions by leveraging the temporal continuity inherent in the movement patterns of subjects.
\end{itemize}
\begin{itemize}
\item[$\bullet$]GAIL-GRU \cite{kuefler2017imitating}: GAIL-GRU combines Generative Adversarial Imitation Learning (GAIL) with GRU. This approach aims to replicate complex trajectory patterns through adversarial training, learning directly from observed behaviors to predict future movements accurately.
\end{itemize}
\begin{itemize}
\item[$\bullet$]CS-LSTM \cite{deo2018convolutional}: CS-LSTM enhances the traditional LSTM framework by incorporating convolutional social pooling layers. This addition allows the model to consider not only individual movement patterns but also the social interactions among multiple agents, making it particularly useful in crowded scenarios.
\end{itemize}
\begin{itemize}
\item[$\bullet$] MATF-GAN \cite{zhao2019multi}: MATF-GAN is designed for complex multi-agent environments, encoding historical trajectories and contextual information into a multi-agent tensor. The model employs convolutional layers to extract interaction dynamics and utilizes an adversarial training approach to refine prediction accuracy.
\end{itemize}
\begin{itemize}
\item[$\bullet$]TS-GAN \cite{wang2020multi}: TS-GAN integrates a social convolution module and a social recurrent module within a generative adversarial framework to model the spatial-temporal relationships among multiple agents. This sophisticated approach enables the prediction of realistic multi-agent trajectories by learning from the nuanced interactions in shared environments.
\end{itemize}
\begin{itemize}
\item[$\bullet$] EquiDiff \cite{chen2023equidiff}: EquiDiff develops a conditional diffusion model with an SO(2)-equivariant transformer to generate future trajectories, which integrates historical data and social interactions through Recurrent Neural Networks and Graph Attention Networks.
\end{itemize}
\subsection{Model Performance Evaluation}
The numerical results of proposed Crossfusor and baseline models in feet are compared in Table \uppercase\expandafter{\romannumeral2} and Figure 6.
\begin{table}[ht]
  \centering
  \caption{RMSE Comparison of Different Models}
   \begin{tabular}{@{}c@{\hskip 28pt}c@{\hskip 26pt}c@{\hskip 26pt}c@{\hskip 26pt}c@{\hskip 26pt}c@{}}
    \toprule[1.0pt]
    Model & 1s & 2s & 3s & 4s & 5s \\
    \midrule
    CV & 2.40 & 5.84 & 10.27 & 15.68 & 21.92 \\
    V-LSTM & 2.23 & 5.41 & 9.55 & 14.63 & 20.57 \\
    GAIL-GRU & 2.26 & 4.95 & 8.37 & 11.98 & 15.45 \\
    CS-LSTM & 2.00 & 4.17 & 6.86 & 10.17 & 14.34 \\
    MATF-GAN & 2.17 & 4.40 & 6.82 & 9.74 & 13.55 \\
    TS-GAN & 1.97 & 4.07 & 6.40 & 9.12 & 12.20 \\
    EquiDiff & \textbf{1.80} & 3.97 & 6.30 & 9.94 & 13.16 \\
    \textbf{Crossfusor} & 2.43 & \textbf{3.64} & \textbf{5.02} & \textbf{6.50} & \textbf{7.97} \\
    \bottomrule[1.0pt]
  \end{tabular}
\end{table}
\begin{figure}[htbp]
\centering
\includegraphics[width=9cm]{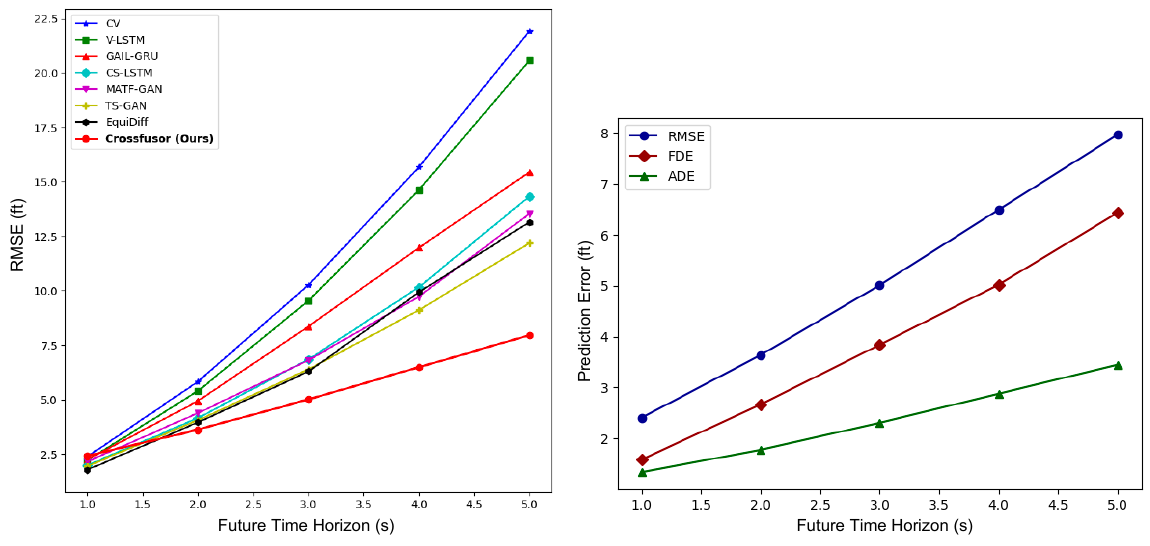}
\caption{Prediction Results of Different Models (Left) and Performance According to Different Metrics (Right)}
\label{fig:yourlabel}
\end{figure}

The experimental results clearly demonstrate the effectiveness of the Crossfusor model in trajectory prediction across various prediction time horizons. Crossfusor showcases an exemplary balance in maintaining lower RMSE values across increasing time intervals, particularly noticeable at longer horizons such as 4s and 5s, where its RMSE remains significantly below those of relatively earlier models like CV and V-LSTM, and even outperforms newer models like MATF-GAN, TS-GAN, and EquiDiff, as illustrated in the left part of Figure 6. 

The right part of Figure 6 further validates the superiority of the Crossfusor model by examining the prediction performance over different time horizons through RMSE, FDE, and ADE metrics. As depicted, Crossfusor maintains consistently lower error values. This trend is particularly evident as the prediction horizon extends to 5 seconds, where the model's performance advantage becomes more pronounced. The lower RMSE, FDE, and ADE values signify the model's accuracy and reliability in long-term trajectory predictions.

The remarkable performance of Crossfusor validates the necessity to address the dynamic complexities of vehicular interactions in car-following scenarios. It suggests that the use of noise scaled by extracted historical information in the forward diffusion process, and the leverage of denoising process guided by the thoroughly encoded car-following interactions are critical for achieving higher accuracy in trajectory prediction.

\subsection{Inference Visualization}
To better illustrate the working process of Crossfusor, we randomly select two distinct platoon examples for demonstration. These examples visualize the reverse denoising process during the inference phase of the Crossfusor model. As shown in Figure 7, each example includes trajectory plots and heatmaps showing the progression of noise reduction and trajectory refinement through successive steps in the denoising process.
\begin{figure*}[htbp]
\centering
\includegraphics[width=18.5cm]{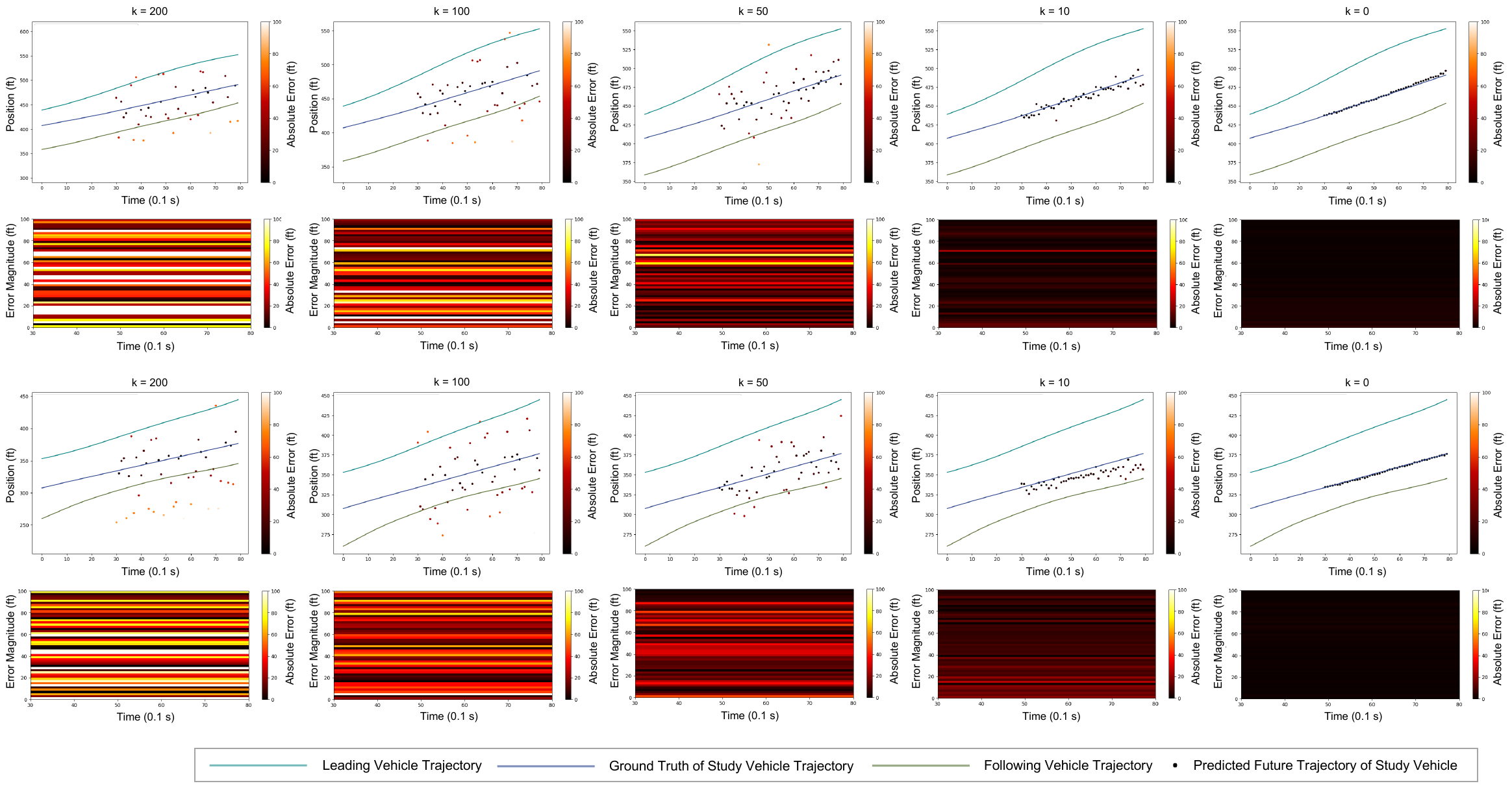}
\caption{Visualization of Denoising Process}
\label{fig:yourlabel}
\end{figure*}

The trajectory plots display the movement of the study vehicle along with the leading and following vehicles over time. As $k$ decreases, which represents a step-by-step reduction of noise in the denoising process, there is a noticeable convergence of the predicted trajectory towards the ground truth trajectory. 
Accompanying each trajectory plot is a heatmap that visualizes the intensity of noise in the system at each corresponding time step. At $k=200$, the denoising process starts with a Gaussian noise scaled by the study vehicle's extracted historical features. The orange and yellows colors in the heatmaps indicate the high noise levels. As the process unfolds, the heatmap colors shift towards red and eventually black, indicating a substantial reduction in noise. By $k=0$, the heatmap shows minimal noise, correlating with the trajectory plot where the predicted trajectory aligns closely with the actual trajectory.

Subsequently, Figure 8 presents multiple scenarios demonstrating the Crossfusor model's robust performance in trajectory prediction across diverse traffic situations. 
\begin{figure*}[htbp]
\centering
\includegraphics[width=16cm]{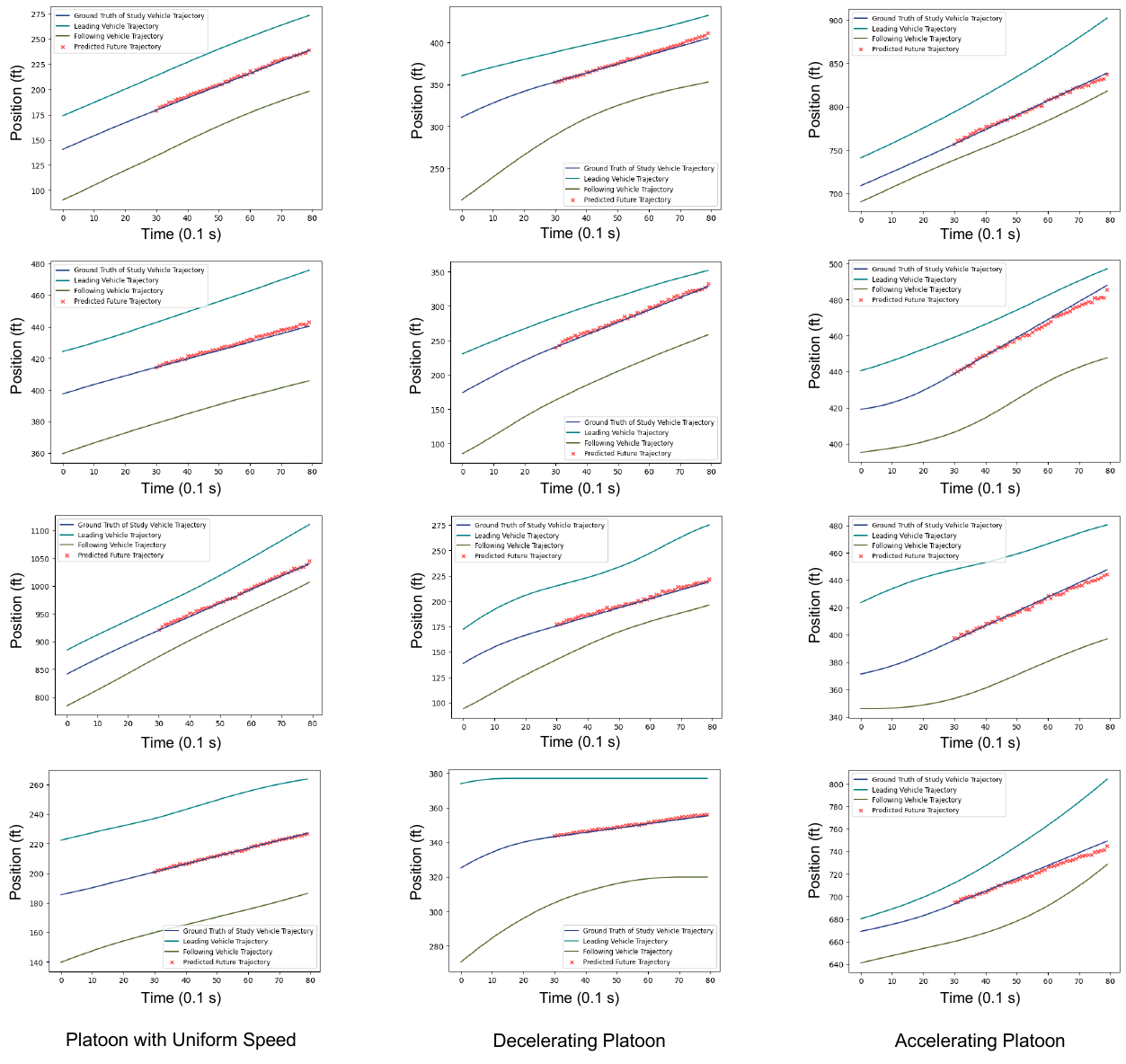}
\caption{Prediction Results Across Different Traffic Situations}
\label{fig:yourlabel}
\end{figure*}
Notably, the model consistently aligns the predicted trajectories (red dots) closely with the ground truth (blue lines), showcasing its accuracy and reliability. This consistency is evident even in complex scenarios where the leading (green lines) and following (black lines) vehicles exhibit non-linear movements or abrupt changes. Such dynamics highlight the model's capability to dynamically incorporate the behavior of surrounding vehicles into its predictions. 

Furthermore, the plots illustrate the model's precision over time and its adaptability to real-time changes in the driving environment. The interaction between the study vehicle and its neighbors significantly influences the trajectory predictions, emphasizing the model's sophisticated use of encoded car-following interactions. These interactions ensure that the predictions adjust based on both historical data and the spatial-temporal context provided by nearby vehicles. 

\subsection{Ablation Study}
To evaluate the contribution of each component in the proposed Crossfusor model, an ablation study is conducted. We construct three variants of the Crossfusor model by systematically removing or altering key components and assessing their impact on prediction performance across various time horizons. The logic of constructing these variants is described as follows:
\begin{itemize}
\item \textbf{Crossfusor w/o Noise Scaling:} This variant removes the noise scaled by the encoded historical trajectory features in the forward process, instead using isotropic Gaussian noise.

\item \textbf{Crossfusor w/o Location Attention and FFT-Based Historical Information Encoding:} This variant eliminates the GRU layers, location-based attention mechanism and the FFT layer used for historical information encoding.

\item \textbf{Crossfusor w/o Cross-Attention Transformer:} This variant replaces the cross-attention transformer mechanism with a linear layer for inter-vehicle dependencies modeling.
\end{itemize}

The result of ablation study is presented in Table \uppercase\expandafter{\romannumeral3}. 
\begin{table*}[ht]
  \centering
  \caption{Result of Ablation Study}
  \begin{tabular}{@{}c@{\hskip 30pt}c@{\hskip 28pt}c@{\hskip 28pt}c@{\hskip 28pt}c@{\hskip 28pt}c@{}}
    \toprule[1.0pt]
    Model & 1s & 2s & 3s & 4s & 5s \\
    \midrule
    Crossfusor w/o Noise Scaling & 2.92 & 3.94 & 5.22 & 6.66 & 8.20 \\
    Crossfusor w/o Location Attention and FFT-Based Historical Feature Encoding & 5.48 & 5.97 & 6.59 & 7.19 & 8.23 \\
    Crossfusor w/o Cross-Attention Transformer & 2.82 & 4.07 & 5.58 & 7.19 & 8.86 \\
    \textbf{Crossfusor} & \textbf{2.43} & \textbf{3.64} & \textbf{5.02} & \textbf{6.50} & \textbf{7.97} \\
    \bottomrule[1.0pt]
  \end{tabular}
\end{table*}
The result highlights the importance of each component in Crossfusor. The variant without noise scaling shows increased RMSE values, particularly over longer time horizons, demonstrating that the use of noise scaled by historical features significantly enhances prediction accuracy and stability. The variant lacking location-based attention and FFT for historical encoding also exhibits higher RMSE values, emphasizing the necessity of these mechanisms for capturing temporal dependencies and spatial relationships in vehicle trajectories. Additionally, the exclusion of the cross-attention transformer results in a notable performance drop, especially for longer predictions, underscoring its critical role in modeling detailed inter-vehicular interactions. Overall, the full Crossfusor model, which integrates noise scaling, the special historical feature extraction pipeline, and the cross-attention transformer, consistently outperforms its variants, illustrating the synergistic effect of these components in improving trajectory prediction accuracy.
\section{Conclusion}
This study presents Crossfusor, a novel Cross-Attention Transformer Enhanced Conditional Diffusion Model for car-following trajectory prediction. Crossfusor integrates noise scaling by encoded historical features and a cross-attention transformer to accurately model detailed inter-vehicular interactions and car-following dynamics. Experimental results on the NGSIM dataset demonstrate that Crossfusor outperforms state-of-the-art models, particularly in long-term predictions, highlighting its robustness and reliability in diverse traffic scenarios. The ablation study further underscores the importance of each component within Crossfusor, showcasing the synergistic effect of integrating each component to enhance prediction accuracy and realism. The inclusion of detailed car-following behaviors and dynamic inter-vehicle interactions ensures a comprehensive and realistic prediction framework, crucial for the advancement of autonomous driving technologies.

While Crossfusor demonstrates significant advancements in trajectory prediction, several avenues for future research remain. First of all, the model can be extended to handle more complex traffic scenarios involving multiple lanes, varying traffic densities, and diverse driving behaviors. Enabling multi-modal prediction capabilities by integrating various types of input data such as vehicle sensor data, data from roadside units (RSUs), and environmental sensing data will enhance the model's ability to predict trajectories under a wider range of conditions and scenarios. Additionally, exploring how the trajectory predictions generated by Crossfusor can be integrated into decision-making and vehicle control strategies is crucial. Investigating the real-time application of these predictions in adaptive cruise control, collision avoidance, and route planning will be essential for advancing autonomous driving technologies.





\ifCLASSOPTIONcaptionsoff
  \newpage
\fi



%


\small
\bibliographystyle{IEEEtran}
\bibliography{IEEEabrv,reference}


\begin{IEEEbiography}
[{\includegraphics[width=1in,height=1.25in,clip,keepaspectratio]{Junwei.jpg}}]{Junwei You}
received the M.S. degree in Civil and Environmental Engineering from Northwestern University in 2022. He is currently a Ph.D. student in Civil and Environmental Engineering at University of Wisconsin-Madison. His research interests include multimodal generative AI, connected and automated vehicles, and deep learning advancement in intelligent transportation systems.
\end{IEEEbiography}

\begin{IEEEbiography}
[{\includegraphics[width=1in,height=1.25in,clip,keepaspectratio]{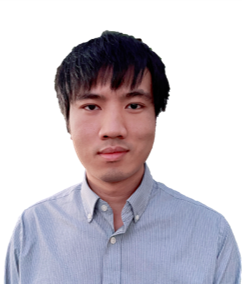}}]{Haotian Shi}
serves as a research associate at University of Wisconsin-Madison. He received his Ph.D. degree in Civil and Environmental Engineering from the University of Wisconsin-Madison in May 2023. He also received three M.S. degrees in Power and Machinery Engineering (Tianjin University, 2020), Civil and Environmental Engineering (UW-Madison, 2020), and Computer Sciences (UW-Madison, 2022). His main research directions are connected and automated vehicles, intelligent transportation systems, traffic crash data analysis, and deep reinforcement learning.
\end{IEEEbiography}

\begin{IEEEbiography}[{\includegraphics[width=1in,height=1.25in,clip,keepaspectratio]{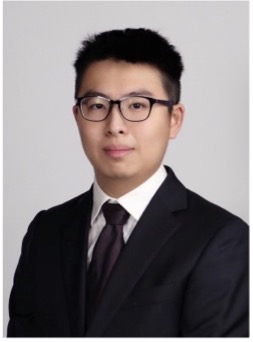}}]{Keshu Wu} 
is currently a postdoctoral research associate at Texas A\&M University. He receives his Ph.D. in Civil and Environmental Engineering from the University of Wisconsin-Madison in 2024. He also holds an M.S. degree in Civil and Environmental Engineering from Carnegie Mellon University in 2018 and an M.S. degree in Computer Sciences from the University of Wisconsin-Madison in 2022. He completed his B.S. in Civil Engineering at Southeast University in Nanjing, China in 2017. His research interests include the application and innovation of artificial intelligence and deep learning techniques in connected automated driving, intelligent transportation systems, and digital twin modeling and simulation.
\end{IEEEbiography}

\begin{IEEEbiography}
[{\includegraphics[width=1in,height=1.25in,clip,keepaspectratio]{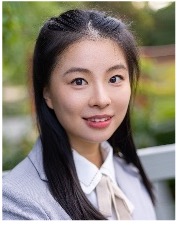}}]{Keke Long}
is a PhD candidate in the Department of Civil and Environmental Engineering at University of Wisconsin, Madison, Madison, WI, USA. She obtained her Master's degree in Traffic Engineering from Tongji University, Shanghai, China in 2021 and B.S. degree in Traffic Engineering from Changan University, Xi’an, China in 2018. Her main research interests are connected and autonomous vehicles and artificial intelligence application in transportation.
\end{IEEEbiography}

\begin{IEEEbiography}
[{\includegraphics[width=1in,height=1.25in,clip,keepaspectratio]{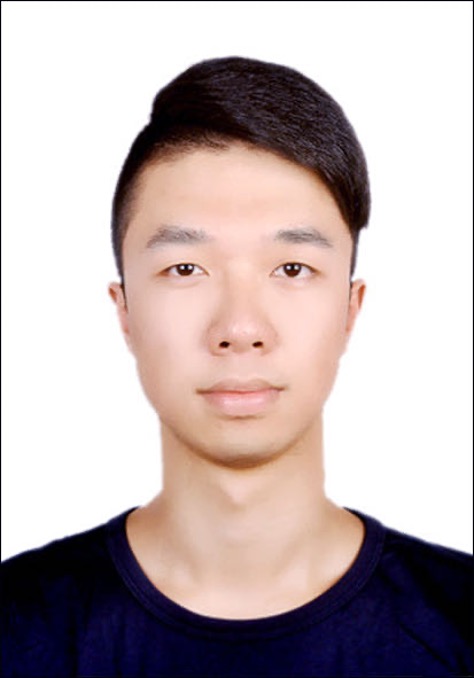}}]{Sicheng Fu}
received the M.S. degree in Civil and Environmental Engineering and Computer Science from the University of Wisconsin Madison, WI, USA, in 2020 and 2022. He is currently pursuing a Ph.D. degree in Civil and Environmental Engineering, at the University of Wisconsin Madison, WI, USA. His main research directions are connected automated vehicles, crash data analysis, and traffic flow data prediction.
\end{IEEEbiography}

\begin{IEEEbiography}
[{\includegraphics[width=1in,height=1.25in,clip,keepaspectratio]{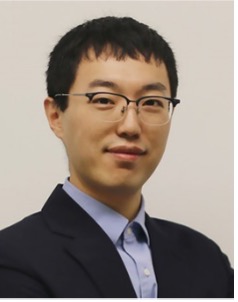}}]{Sikai (Sky) Chen}
is an Assistant Professor at the Department of Civil and Environmental Engineering and the Department of Mechanical Engineering (courtesy), University of Wisconsin-Madison. He received his Ph.D. in Civil Engineering with a focus on Computational Science \& Engineering from Purdue University in 2019. His research centers around three major themes: human users, AI, and transportation. He aims to innovate and develop safe, efficient, sustainable, and human-centered transportation systems using cutting-edge methods and technologies. The focus is on incorporating human behaviors, interactive autonomy, digital infrastructure, and intelligent control frameworks. In addition, he is a member of two ASCE national committees: Connected \& Autonomous Vehicle Impacts, and Economics \& Finance; IEEE Emerging Transportation Technology Testing Technical Committee, and TRB Standing Committee on Statistical Methods (AED60). More information can be found at Sky-Lab: https://sky-lab-uw.github.io/.
\end{IEEEbiography}

\begin{IEEEbiography}
[{\includegraphics[width=1in,height=1.25in,clip,keepaspectratio]{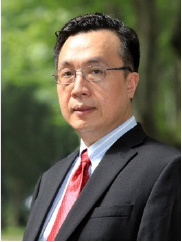}}]{Bin Ran}
is the Vilas Distinguished Achievement Professor and Director of ITS Program at the University of Wisconsin at Madison. Dr. Ran is an expert in dynamic transportation network models, traffic simulation and control, traffic information system, Internet of Mobility, Connected Automated Vehicle Highway (CAVH) System. He has led the development and deployment of various traffic information systems and the demonstration of CAVH systems. Dr. Ran is the author of two leading textbooks on dynamic traffic networks. He has co-authored more than 240 journal papers and more than 260 referenced papers at national and international conferences. He holds more than 20 patents of CAVH in the US and other countries. He is an associate editor of Journal of Intelligent Transportation Systems.
\end{IEEEbiography}

\end{document}